\documentclass{IEEEtran}
\usepackage{cite}
\usepackage{amsmath,amssymb,amsfonts}
\usepackage{algorithmic}
\usepackage{graphicx}
\usepackage{textcomp}
\usepackage{subfigure}
\usepackage[none]{hyphenat}

\usepackage{hyperref}       
\usepackage{url}            
\usepackage{booktabs}       
\usepackage{amsfonts,amsmath}       
\usepackage{nicefrac}       
\usepackage{microtype}      
\usepackage{graphicx}
\usepackage[square,sort,comma,numbers]{natbib}
\usepackage{doi}

\begin{document}

\title{Predicting Gender by First Name Using Character-level Machine Learning}
\author{\uppercase{Rosana C B Rego}, 
\uppercase{Verônica M L Silva and Victor M Fernandes}
}


\maketitle

\begin{abstract}
 Predicting gender by the first name is not a simple task. In many applications, especially in the natural language processing (NLP) field, this task may be necessary, mainly when considering foreign names.  In this paper, we examined and implemented several machine learning algorithms, such as extra trees, KNN, Naive Bayes, SVM, random forest, gradient boosting, light GBM, logistic regression, ridge classifier, and deep neural network models, such as MLP, RNN, GRU, CNN, and BiLSTM, to classify gender through the first name. A dataset of Brazilian names is used to train and evaluate the models. We analyzed the accuracy, recall, precision, f1 score, and confusion matrix to measure the models' performances. The results indicate that the gender prediction can be performed from the feature extraction strategy looking at the names as a set of strings. Some models accurately predict gender in more than $95\%$  of the cases. The recurrent models overcome the feedforward models in this binary classification problem. 

\end{abstract}

\textit{Keywords:
Machine learning, Deep Learning,  Gender Prediction, Natural language processing.
}

\section{Introduction}
\label{sec:introduction}
Prediction problem refers to a wide class of problems in which the goal is to perform an inference based on situational plays and statistical-based models. However, in some situation is not easy to make an inference. In this way, artificial intelligence (AI) algorithms can be applied to produce a prediction in which the problem goal is to provide the correct label (e.g., prediction or output) to an instance (e.g., set of features or inputs). One of the well-known prediction problems is gender inference.  

In some applications,  such as investigations of psychological, anthropological, and sociological research questions \cite{karimi2016inferring}, inferring the gender of a person is necessary. But how could we infer gender? Indeed there are many research efforts all over the world in this field that are centered in many directions, such as natural processing language (NLP) \cite{vicente2019gender,al2019study}, computer vision \cite{ugail2018secrets,galla2020support}, and image processing \cite{afifi2019afif4}. 

The completeness of the database used in training and inference models and algorithms is essential for applications of AI. Sometimes the dataset does not include an attribute that is important to the problem, such as the lack of the person's gender. To solve this problem, it is possible to use pre-existing information to predict other information needed to fill the database. Using some basic logic and machine learning, we can infer gender by taking the name and compares it against a list of names associated with either a male or female gender, as explained in the following papers \cite{panchenko2014detecting,van2020gender,hu2021s}.

Gender prediction from other attributes is a technique already used in artificial intelligence applications. Many works predict gender from images of people's faces, such as \cite{Age2017, transfer2020, SVM2020, Census2020, predict2019, surinta2019}. In the paper \cite{Age2017}, the authors studied the deep residual networks of residual networks (RoR) capability to achieve gender prediction by analyzing images. In the same way, Wang et al. \cite{transfer2020} explored the deep convolutional neural networks (CNN) capability to perform gender recognition by examining images. Venugopal et al. \cite{SVM2020} applied the support vector machine (SVM) algorithm to classify children's gender from images. Also, using SVM, Kuehlkamp and Bowyer \cite{predict2019} predicted gender by processing images. Furthermore, the authors compared SVM with the CNN network. In \cite{surinta2019}, the authors explored the image processing techniques through an SVM classifier. 

Other researchers proposed gender prediction using natural language processing \cite{Vashisth2020, Lekamge2019, Mamgain2019}. For instance, Vashisth and Meehan \cite{Vashisth2020} explored some NLP techniques, such as bag of words, word embedding, logistic regression, SVM, and Naive Bayes, to infer the gender of a person based on Twitter data. In the work \cite{Lekamge2019}, the authors proposed two Long Short Term Memory (LSTM) to predict the effect of Gana on personal names and perform gender inference.  Mamgai et al. \cite{Mamgain2019} applied logistic regression, random forest, bag-of-words, and LSTM-CNN to infer gender and language variety of authors. They concluded that the LSTM-CNN model works better for language variety tasks and bag-of-words gives good accuracy for gender prediction. Motivated by the previous discussion, we proposed some character-level deep learning models based on NLP to infer a person's gender by the first name. 

Deep learning is a machine learning subfield and it has been widely applied to prediction problems \cite{Otter2021}. Deep learning, also known as deep neural network, can be classified into deep feedforward neural networks (DFNN) or just multilayer perceptron (MLP), CNN, and recurrent and recursive neural networks (RNN) \cite{goodfellow2016deep}. 

Based on the capability of the deep models, we implemented five character-level deep neural network models as MLP, CNN, RNN, bidirectional long short term memory (BiLSTM), and gated recurrent unit (GRU) to realize the gender prediction through the first name. Moreover, to compare the performance of deep learning models with machine learning models, we implemented thirteen character-level machine learning algorithms, as extra trees classifier, k-nearest neighbors (KNN), Naive Bayes, Support vector machine (SVM), random forest, gradient boosting, light gradient boosting (LightGBM), logistic regression, ridge classifier, decision tree, Ada boost, linear discriminant analysis (LDA), and quadratic Discriminant Analysis (QDA).
Thus, the main contributions of this work are:
\begin{itemize}
    \item[1.] Pre-processing names using character encoding via one-hot encoding technique and predicted gender with machine learning and deep learning models based on the person's first name.
    
    \item[2.] Analysis of the most common machine learning algorithms applied to the binary classification problem as well as performance comparison with deep neural networks models. 
    
    \item[3.] Comparison of the two categories of deep learning models, such as feedforward (MLP, CNN) and recurrent (RNN, GRU, BiLSTM) networks for binary classification problems and model performance evaluation through accuracy, recall, precision, and confusion matrix. 
 \end{itemize}
 
 The paper is divided as follows: In section 2, some related works are presented. In section 3, the dataset, the pre-processing procedures adopted and the evaluation metrics are described. In section 4, the machine learning algorithms and deep learning models are introduced.  In section 5, an experimental simulation is performed. Finally, in section 6, the final considerations of the study are presented.

\section{Related Work}

There has been some research that uses the name of a person to infer gender \cite{panchenko2014detecting,Amarappa2015Kannada,manik2019gender,van2020gender,yuenyong2020gender,hu2021s}. Panchenko and Teterin \cite{panchenko2014detecting} first proposed gender prediction using the full name of a person. They consider Russian full names and used a 2-regularized Logistic Regression model for inference, which given an accuracy up to 96\%. In the work \cite{Amarappa2015Kannada} a statistical machine learning to named entity recognition and classification system for the Kannada Language is proposed. The system includes identification of proper names in texts and classification of those names into a set of categories such as person names and organization names. In addition, a Naive Bayes model is evaluated for Kannada Indian names resulting in a 10-fold cross-validation accuracy of 77.2\%. 
  
A logistic regression model is applied on Indonesian names for gender inference in \cite{manik2019gender} resulting in an accuracy of 83.7\%. Logistic regression is further combined with a CNN modeled to infer gender through profile photos resulting in an improvement in accuracy to 98.6\%. Also, in \cite{van2020gender}, a logistic regression model is applied to predict the gender of written and spoken Chinese names.  The authors verified that having both modalities available (i. e., written and spoken) did not significantly increase the identification of a name as female or male compared to the written-only condition.

Machine learning models such as SVM, Naive Bayes, Random Forest are used in  \cite{yuenyong2020gender} to identify gender through the first name and nickname of Thai Facebook profiles. The results obtained using word tokenization present an accuracy of 65\% for the base models and more than 90\% for the combined models.

In \cite{hu2021s}, character-based machine learning models are compared with models based on content information for gender prediction. Character-based models performed better for the trained data.


In this work, the best machine learning and deep learning models found in the related works previously stated as well as a set of other models are tested with Brazilian names to perform a systematic comparison between them. We also included other metrics beyond accuracy, such as recall, precision and F1 score.

\section{Dataset and Pre-processing}
 
The dataset and the pre-processing techniques used are described in this section.

\subsection{Dataset}
\begin{figure}[!htb]

\subfigure[]{\includegraphics[width=0.375\textwidth]{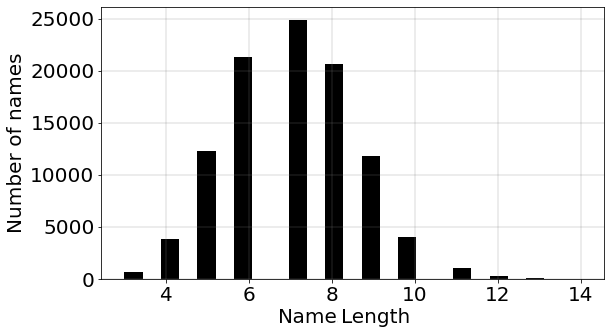}}
\subfigure[]{\includegraphics[width=0.4\textwidth]{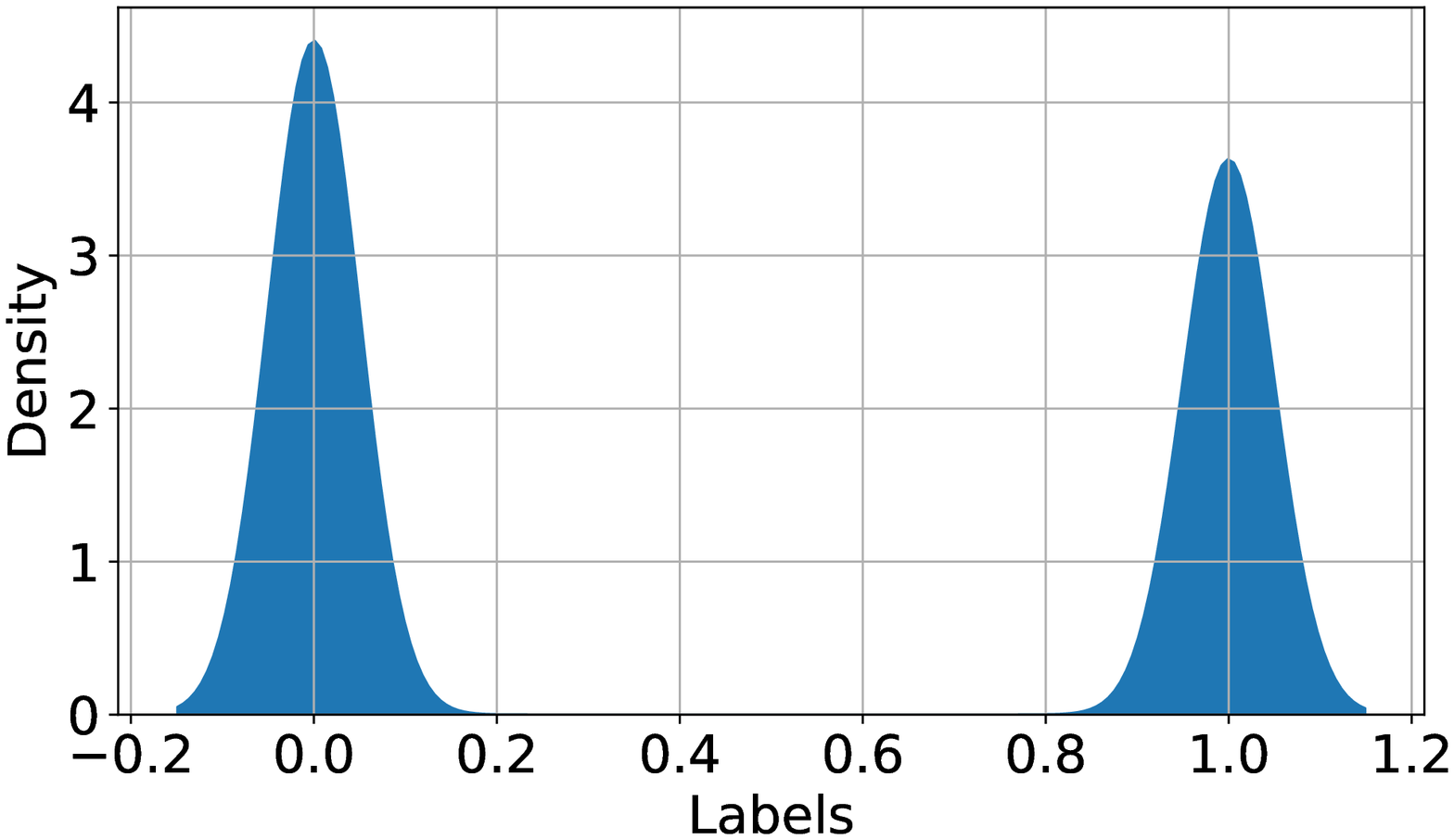}}

\caption{(a) Name length and (b) Labels distribution.}
\label{lennames}
\end{figure}

The dataset consists of 100787 Brazilian names, of which 54.82$\%$ are female names and 45.18$\%$ are male, based on 2010 CENSO data from Brasil.io\footnote{https://brasil.io/dataset/genero-nomes/grupos/} project. The name length distribution is depicted in Figure \ref{lennames} (a). The labels distribution considering the gender (0 for female and 1 for male) is depicted in Figure  \ref{lennames} (b). We are particularly examining the first name of a person. Consequently, when we talk about a name, we are referring to the first name. Into the dataset, we have gender information, name, total frequency, group frequency, group name, and ratio. The first ten rows of the dataset are depicted in Table \ref{tabnames}.

\begin{table}[!htb]
\caption{Dataset's first 10 rows. $Total_f$ denotes total frequency and $Group_f$ the group frequency.}
\label{tabnames}
\begin{tabular}{|p{22pt}|l|p{20pt}|p{22pt}|p{29pt}|l|l}
\hline
Gender & Name & Total$_f$ & Group$_f$ & Group name & Ratio\\ \hline
M & AARAO &   281 & 3526 & ARAO & 1.0 \\
M & AARON &   676 & 3442 & ARON & 1.0 \\
F & ABA &  82 & 5583 & ADA & 1.0 \\
M & ABADE &   57 & 57 & ABADE & 1.0 \\
M & ABADI &  73 & 116 & ABADI & 1.0 \\
F & ABADIA &  7565 & 7565 & ABADIA & 0.9832 \\
M & ABADIAS &  201 & 201 & ABADIAS & 0.7761 \\
 M & ABADIO & 1550 & 1550 & ABADIO & 1.0 \\
M & ABAETE &   39 & 233 & ADETE & 1.0 \\
 M & ABD &   23 & 23 & ABD & 1.0 \\
\hline
\end{tabular}
\end{table}

Note that in the dataset, we have a gender ratio for each name, that is because some names are found in both genders. With this in mind, we prepared two subsets of data, one using only names with 100\% gender ratio and the other with all data. The dataset with 100\% gender ratio consists of 90158 Brazilian names, in which 40303 are male and 49855 are female. 

\subsection{Word Encoding}

Machine learning algorithms, such as neural network can only discover patterns in numerical data, so it is needed to transform our data into numeric values with the process called encoding word. The word encoding process consists of converting the letters into numbers. Then, the first step to encode the letter is to define a glossary that corresponds to all the single letters encountered in the alphabet and maps each letter of the glossary to a number. We used the \textit{word2vec} technique to encode the names, which used one-hot encoding \cite{mikolov2013efficient}. We set a vector with zeros and a 1 which represents the corresponding letter included in the name and existing vocabulary/glossary. Our vocabulary has 28 letters and we set a maximum number of characters by name as 20. 

Therefore, if we have the name "\textit{ana}" as input, after the one-hot encoding, the name will be represented as 

\begin{equation*}
\tiny
  \begin{array}{ll}
        a: & [1, 0, 0, 0, 0, 0, 0, 0, 0, 0, 0, 0, 0, 0, 0, 0, 0, 0, 0, 0, 0, 0, 0, 0, 0, 0, 0, 0]\\
        n: & [0, 0, 0, 0, 0, 0, 0, 0, 0, 0, 0, 0, 0, 1, 0, 0, 0, 0, 0, 0, 0, 0, 0, 0, 0, 0, 0, 0] \\
        a: & [1, 0, 0, 0, 0, 0, 0, 0, 0, 0, 0, 0, 0, 0, 0, 0, 0, 0, 0, 0, 0, 0, 0, 0, 0, 0, 0, 0]
        .\end{array} 
\end{equation*}

With this encoding process, our network models will be able to predict accurately the gender of names, as we showed in the numerical simulation section. 

\section{Character-level machine learning classifiers}

\begin{figure}[!htb]
    \centering
    \includegraphics[width=0.4\textwidth]{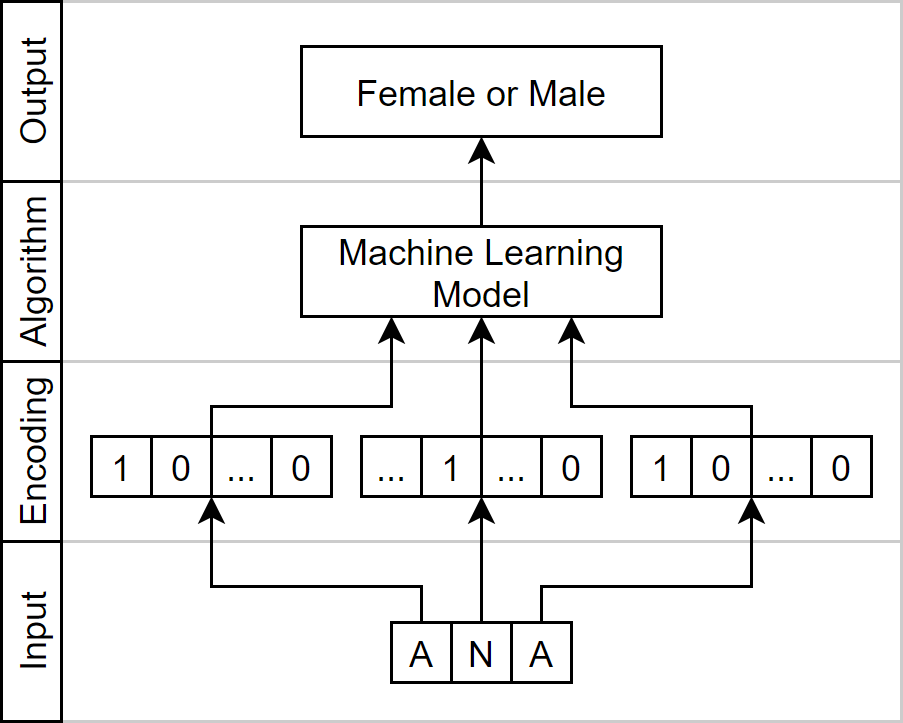}
    \caption{Character level machine learning classification.}
    \label{mg}
\end{figure}

Figure \ref{mg} shows how the machine learning algorithms are applied to learn and make predictions using names. We implemented thirteen character-level machine learning algorithms: the extra trees and decision tree classifier, KNN, Naive Bayes, SVM, random forest, gradient boosting and light gradient boosting, logistic regression, ridge classifier,  Ada boost, LDA, and QDA.

\subsection{Decision trees}

The decision trees algorithm is proposed by Quinlan \cite{quinlan1986induction} and it works for both classification \cite{potharst2000decision} and regression [X] problems. As our problem is a classification problem, this algorithm suits well. We created a decision tree model that predicts the gender of a name considering the Gini impurity function. We considered the whole training set as the root. Based on gender, the character-level vector is distributed recursively. As the decision tree algorithm usually overfits, there are some improved algorithms based on the decision tree, such as random forest and extra trees that will be explained in the following subsections.

\subsubsection{Random forest classifier}
The random forest algorithm decreases the risk of overfitting by building multiple trees and drawing conclusions with replacement \cite{ahmad2018performance}. We implemented a random forest classifier with 100 as the number of trees in the forest. Moreover, we used the Gini impurity function to measure the quality of a split. 

\subsubsection{Extra trees}
Extra trees algorithm is like the random forest, but it does not draw observations with replacement. Moreover, the tree nodes are split based on random splits \cite{alsariera2020ai}. So, we also implemented an extra-trees classifier with the number of trees in the forest as 100, and the function to measure the quality of a split as Gini impurity. Moreover, we used the whole dataset to build each tree. 

\subsection{Boosting Algorithms: Gradient, LightGBM and Ada}

Boosting algorithms are based on a tree ensemble algorithm. In the boosting algorithms, the trees are grown sequentially and using the information from the previous tree \cite{alzamzami2020light}. We implemented three boosting algorithms: Gradient, LightGBM, and Ada boosting. 

The gradient boosting used only a single regression tree. We applied the deviance or logistic regression as loss function. For the LightGBM model, we used the number of boosted trees to fit as 100 and  the maximum tree leaves for base learners as 31. For Ada boosting, we used the maximum number of estimators at which boosting is terminated as 50 and the weight applied to each classifier at each boosting iteration as 1.

\subsection{Support vector machine}

Support vector machine can be used for both classification and regression problems \cite{jan2017sensor}. As we discussed previously, it has been applied to perform gender classification. To compare SVM with others machine learning algorithms, we  implemented a support vector machine with a linear kernel. 

\subsection{Naive Bayes}

Naive Bayes is a probabilistic classifier based on the Bayes' theorem \cite{xue2020real}. We applied the categorical naive Bayes to classify the gender of a given name. 

\subsection{KNN}
KNN algorithm was proposed by Hodges et al. \cite{fix1989discriminatory}. KNN is used for both classification and regression \cite{zhao2018knn}. In this paper, we executed a K-nearest neighbors for binary classification, in which the classifier has the number of neighbors as 5 and the uniform weight function.

\subsection{Classification by Regression: Logistic and Ridge}

We constructed a classification model using a regression learner. The classification models are built from logistic regression and Ridge regression. The logistic regression classifier was implemented considering tolerance for stopping criteria as 0.0001 and the norm-L2 in the penalization. Our second classifier uses Ridge regression. We set the precision of the solution as 0.001 and we used the conjugate gradient as solver.

\subsection{Linear and quadratic discriminant analysis}

The linear and quadratic discriminant analyses are two well-known classifiers \cite{hong2017comparative} and we implemented both. Each one computes the probability distribution of being classified the name as female or male. Both assume that the names from each gender are drawn from a Gaussian distribution. However, the QDA of each gender has its covariance matrix. 

\subsection{Deep learning models}

All deep learning models need a loss function to compute how well the model is performing during the training. As our problem is a binary classification (1 or 0, M or F), we selected the binary cross-entropy loss function to train each deep learning model. Therefore, the loss function is given by

\begin{equation}
\small
L(p,q) = -\frac{1}{N}\left[\sum^N_{i=1} y_i \cdot log(p(y_i)) + (1-y_i) \cdot log(q(y_i))\right],
\end{equation}
where $y$ is the label (0 for female and 1 for male), $p(y)$ is the predicted probability of the gender being male for all N points, and  $q(y_i)=1-p(y_i)$ is the predicted probability of the gender being female.

We designed five deep neural network models: MLP, CNN, simple RNN, BiLSTM, and GRU. Each of them will be detailed in the following topics.

\subsubsection{Multilayer Perceptrons}

A multilayer perceptron (MLP) or deep feedforward networks learn deterministic mappings from input to output that lack feedback connections \cite{goodfellow2016deep}. An MLP can classify data that is not linearly separable. In natural language processing (NLP), it can be applied to speech recognition and machine translation \cite{kamath2019deep}. We implemented an MLP with two hidden layers. For the first hidden layer with 64 neurons, the relu activation function was considered.  For the second hidden layer with 128 neurons, the softplus activation function was considered. For the output layer, we set the sigmoid activation function. 

\subsubsection{Convolutional Neural Networks}
\begin{figure}[!htb]
    \centering
    \includegraphics[width=0.5\textwidth]{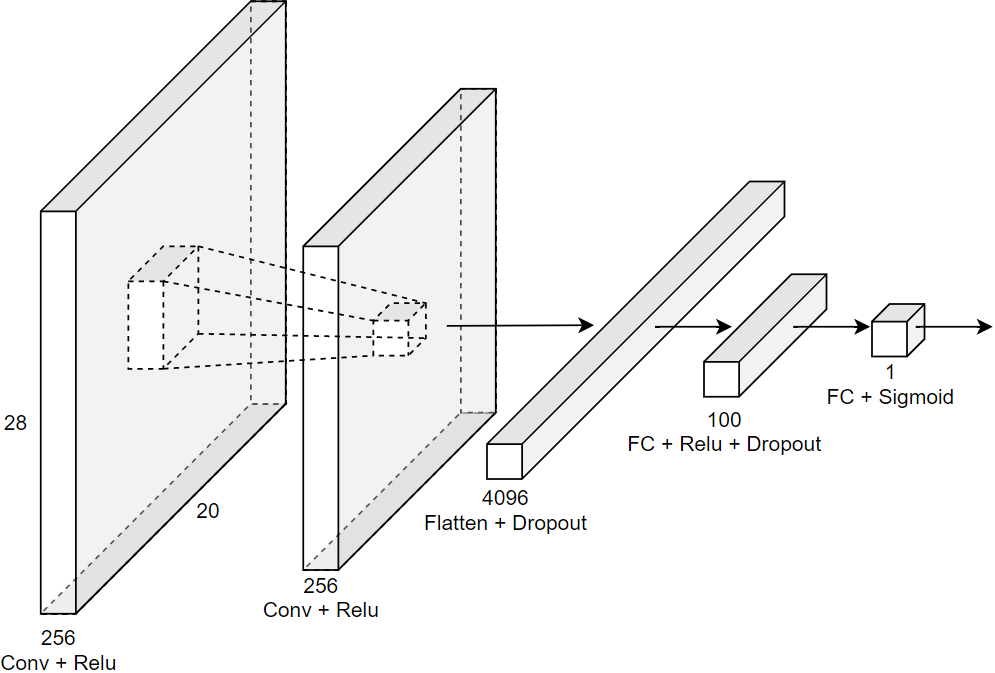}
    \caption{Convolutional neural network model.}
    \label{cnn}
\end{figure}

Character-level CNN (char-CNN) can be used in natural process language as presented by  \cite{zhang2015character}. We proposed a character-level ConvNet composed of two convolutional layers with relu as activation function and two fully-connected layers. The output layer activation function is sigmoid. The suggested char-CNN architecture to gender names classification is displayed in Figure \ref{cnn}. 

\subsubsection{Recurrent Neural Networks}

Basically, there are two functional uses of feedback or recurrent neural networks (RNN): associative memories and input-output mapping networks \cite{haykin2010neural}. There are many architectures in the literature for recurrent networks, such as those based on a recurring input-output model, state-space, recurrent multi-layer perceptron, fully connected recurrent neural network (FCRNN), among others. All these architectures exploit the multi-layer perceptron nonlinear mapping capabilities \cite{haykin2010neural}. However, the RNN output depends not only on the current network input but also on the current or previous network outputs or states. They have important features not found in feedforward networks, such as the ability to store information for later use \cite{goodfellow2016deep}. For this reason, recurrent networks are more powerful than non-recurring networks to process sequential data. 

\begin{figure}[!htb]
    \centering
    \includegraphics[width=0.4\textwidth]{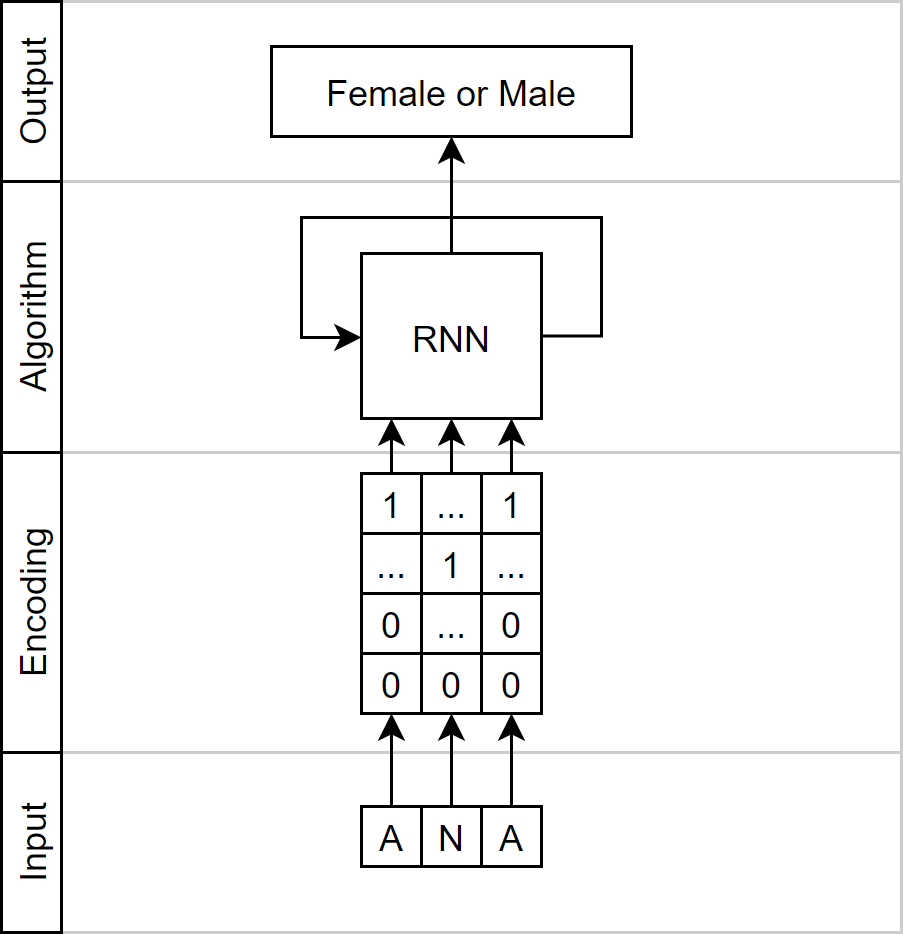}
    \caption{Simple RNN model.}
    \label{rnn}
\end{figure}

We implemented a simple RNN with 32 units, and the sigmoid activation function in the output. As shown in Figure \ref{rnn}, the RNN receives the character sequence as input and returns the gender classification as output.

\subsubsection{Bidirectional Long Short Term Memory Networks}

\begin{figure}[!htb]
    \centering
    \includegraphics[width=0.4\textwidth]{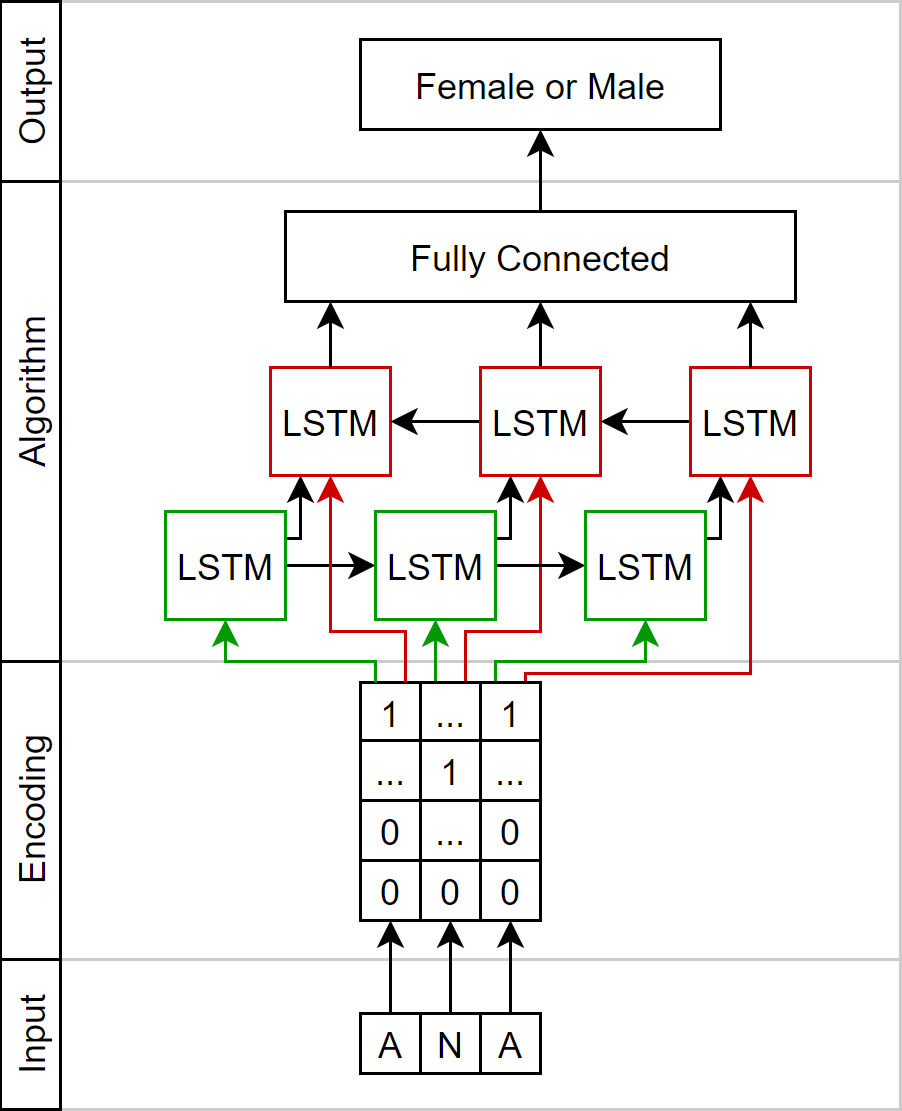}
    \caption{Bidirectional long short term memory model.}
    \label{bilstm}
\end{figure}

A Long Short-Term Memory is a popular RNN proposed by \cite{hochreiter1997long} to solve the problem of vanishing gradients that happens when we are training a simple RNN. A Bidirectional Long Short Term Memory Network is an LSTM modification in which the BiLSTM network scans at a particular sequence both from front to back and from back to front as depicted in Figure \ref{bilstm}.

We designed a BiLSTM model with 64 units in each forward and backward layer. We added a dropout rate of $20\%$ and $L2$ regularization penalty with a factor of $0.002$, and we applied the sigmoid activation function in the output.

\subsubsection{Gated recurrent units}

The Gated recurrent unit neural network proposed by \cite{cho2014learning} has been created to solve the vanishing gradient problem, as LSTM.  Both are designed similarly. To understand better the difference between GRU and LSTM see the paper \cite{chung2014empirical}.

We also built a GRU model with 32 units using the sigmoid activation function in the output to analyze with the BiLSTM model, since both are related. 


\section{Evaluation Metrics}
In order to validate the trained models we performed four evaluation metrics: accuracy, recall, precision and F1 score.

The accuracy is defined by the equation
\begin{equation}
    accuracy:=\frac{TN+TP}{TN+FP+FN+TP},
\end{equation}
where $TP$ means true positive value, $TN$ means true negative, $FP$ means false positive, and $FN$ means false negative value. Because accuracy is not always a sufficient performance measure, we calculated the recall, also known as the true positive rate given by 
\begin{equation}
    recall:=\frac{TP}{FN+TP}.
\end{equation}

Moreover, we also determined the precision defined by 
\begin{equation}
    precision:=\frac{TP}{FP+TP}.
\end{equation}

Through the examination of precision, recall, and accuracy, we can fully evaluate the effectiveness of the proposed models. 

\section{Results}

All models were implemented using python 3.6.9. The deep learning models were built on Keras framework and Tensorflow 2.5. We used two datasets, the first one considering only the names with $100\%$ gender ratio and the second one with all names regardless the gender ratio. Each dataset was splitted into train data, validation data, and test data with $60\%$, $20\%$, $20\%$ of the dataset, respectively.

In the deep neural network models, we used a learning rate (LR) $LR=0.001$, a batch size of $256$, 100 epochs, and an early stopping technique to avoid overfitting. Our problem consists of binary classification, so for all deep models, the binary cross-entropy loss was considered. Moreover, all the deep models were trained with the Adam algorithm.

The accuracy achieved, with the test data, for all models is depicted in Tables \ref{tb} and \ref{t2}. The extra-trees classifier with the number of trees in the forest as 100 performed better than the others machine learning algorithms, except for deep models, such as CNN and BiLSTM, as shown in Tables \ref{tb} and \ref{t2}. 

\begin{table}[!htb]
\caption{Model-performance metrics on the $100\%$ gender ratio dataset.}\label{tb}
\begin{tabular}{|l|l|l|l|l|l|}
\hline
\textbf{Model}                  & \textbf{Accuracy} & \textbf{Recall} & \textbf{Precision} & \textbf{F1} \\
\hline
Extra Trees           & 0.9609                   & 0.9500          & 0.9620         & 0.9560      \\
Random Forest         & 0.9586                   & 0.9453          & 0.9614         & 0.9532      \\
LightGBM & 0.9398                   & 0.9292          & 0.9356         & 0.9324      \\
Decision Tree        & 0.9320                 & 0.9218          & 0.9257         & 0.9237      \\
KNN         & 0.9193                  & 0.8805          & 0.9349         & 0.9069      \\
Logistic Regression             & 0.8889                   & 0.8517          & 0.8944         & 0.8725      \\
SVM             & 0.8863                   & 0.8453          & 0.8960         & 0.8688      \\
LDA    & 0.8817                  & 0.8186          & 0.9074         & 0.8607      \\
Ridge Classifier                & 0.8815                   & 0.8179          & 0.9075         & 0.8604      \\
Gradient Boosting     & 0.8548                  & 0.7260          & 0.9351         & 0.8169      \\
Ada Boost             & 0.8495                  & 0.7516          & 0.8948         & 0.8169      \\
Naive Bayes                     & 0.7237                   & 0.3974          & 0.9611         & 0.5623      \\
QDA & 0.6673                   & 0.2668          & 0.9610         & 0.4153     \\
CNN                             & $0.9686$                     & 0.9569         & 0.9725         & 0.9634 \\
MLP                             & $0.8880$                   & 0.8451         & 0.8986         &   0.8671  \\
BiLSTM                          & $ 0.9696$                   & 0.9597         & 0.9720       &    0.9642 \\
RNN                             & $0.9530$                   & 0.9458         & 0.9490        &    0.9456 \\

GRU                             & $0.9650$                   & 0.9646          & 0.9574        &   0.9594  \\
\hline
\end{tabular}
\end{table}

\begin{table}[]
\caption{Model-performance metrics with all names regardless the gender ratio.}\label{t2}
\begin{tabular}{|l|l|l|l|l|l|}
\hline
\textbf{Model}                  & \textbf{Accuracy}  & \textbf{Recall} & \textbf{Precision} & \textbf{F1} \\
\hline
Extra Trees           & 0.9482                   & 0.9351          & 0.9498         & 0.9424      \\
Random Forest        & 0.9460                   & 0.9311          & 0.9487         & 0.9398      \\
LightGBM & 0.9222                   & 0.9129          & 0.9152         & 0.9140      \\
Decision Tree         & 0.9210                   & 0.9114          & 0.9139         & 0.9126      \\
KNN          & 0.9034                   & 0.8649          & 0.9171         & 0.8902      \\
Logistic Regression             & 0.8672                   & 0.8279          & 0.8725         & 0.8496      \\
SVM              & 0.8661                   & 0.8317          & 0.8684         & 0.8489      \\
Ridge Classifier                & 0.8604                   & 0.7946          & 0.8855         & 0.8375      \\
Gradient Boosting    & 0.8339                   & 0.6864          & 0.9283         & 0.7891      \\
Ada Boost            & 0.8263                   & 0.7335          & 0.8629         & 0.7927      \\
Naive Bayes                     & 0.7076                   & 0.3715          & 0.9559         & 0.5350      \\
LDA    & 0.6886                   & 0.6365          & 0.7083         & 0.6705      \\
QDA & 0.6587                   & 0.2707          & 0.9255         & 0.4168   \\
CNN                             & $0.9578$                  & 0.9494          & 0.9568         &    0.9515  \\
MLP                             & $0.8698$                   & 0.8444          & 0.8642         &   0.8492  \\
BiLSTM                          & $0.9617$                 & 0.9585          & 0.9569         &    0.9564  \\
RNN                             & $0.9400$                   & 0.9336          & 0.9335         &    0.9320  \\
GRU                             & $0.9500$                   & 0.9452          & 0.9442         &    0.9425  \\
\hline
\end{tabular}
\end{table}

\begin{figure}[!htb]
    \centering
\subfigure[]{    \includegraphics[width=1\linewidth]{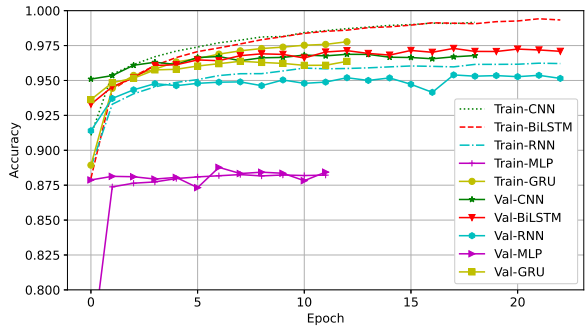}}

\subfigure[]{    \includegraphics[width=1\linewidth]{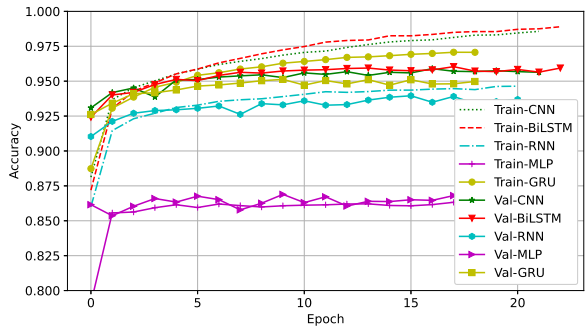}}
\caption{Accuracy versus epochs for deep learning models: (a) Dataset with only $100\%$ gender ratio names and (b) Dataset with all names regardless the gender ratio.}
    \label{figac}
\end{figure}

The accuracy progression through epochs during the train and validation are shown in Figures \ref{figac} (a) and (b). Based on the accuracy, we can infer that the recurrent neural networks perform better to predict the gender for a given name. The deep models achieve accuracy greater than $90\%$, excluding the MLP model, which accomplishes $88.80\%$ and $86.98\%$ with only $100\%$ gender ratio names and regardless the gender ratio, respectively.

\begin{figure}[!htb]
    \centering
\subfigure[]{   \includegraphics[width=1\linewidth]{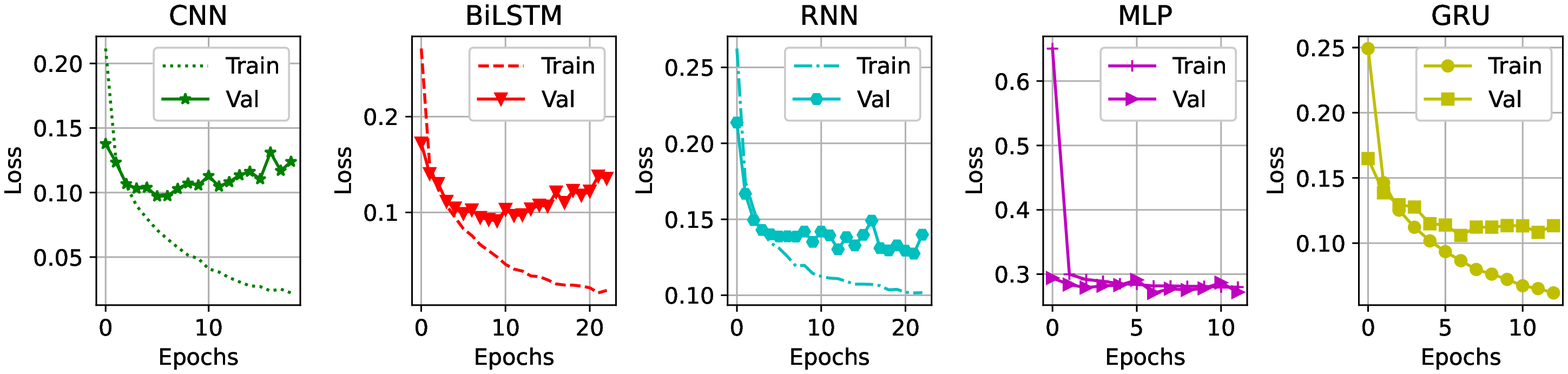}}
 \subfigure[]{   \includegraphics[width=1\linewidth]{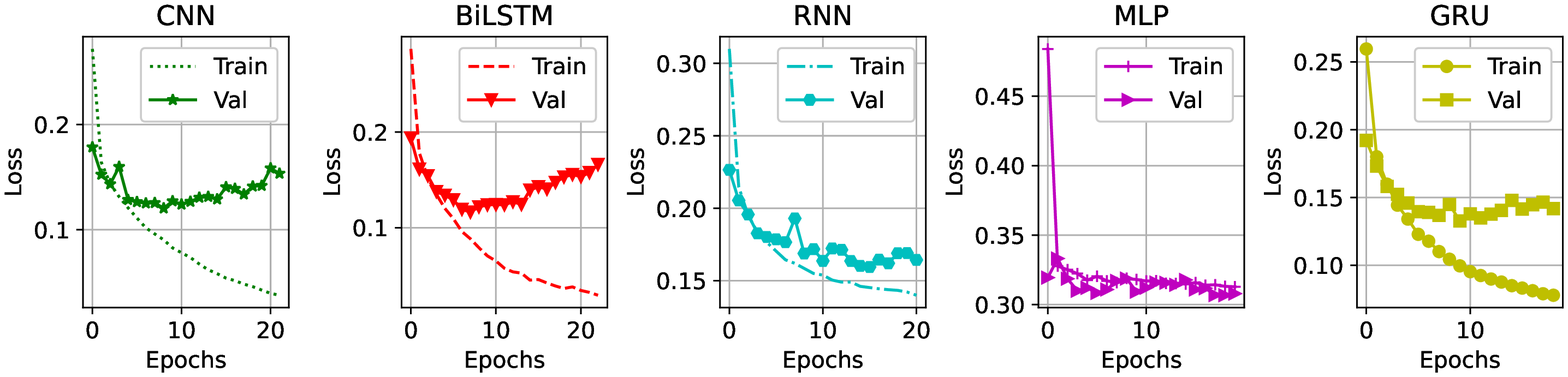}}
    \caption{Loss versus epochs: (a) Dataset with only $100\%$ gender ratio names and (b) Dataset with all names regardless the gender ratio.}
    \label{figloss}
\end{figure}

The values assumed by the loss function, during the train and validation, through epochs are displayed in Figures \ref{figloss} (a) and (b). In the MLP model, the values during the validation are less than the train this is because of the dropouts in our model, it is applied during training, but not during the validation. For all deep models, we set an early stop to avoid the lack of model generalization. 

\begin{figure}[!htb]
    \centering
\subfigure[]{   \includegraphics[width=1\linewidth]{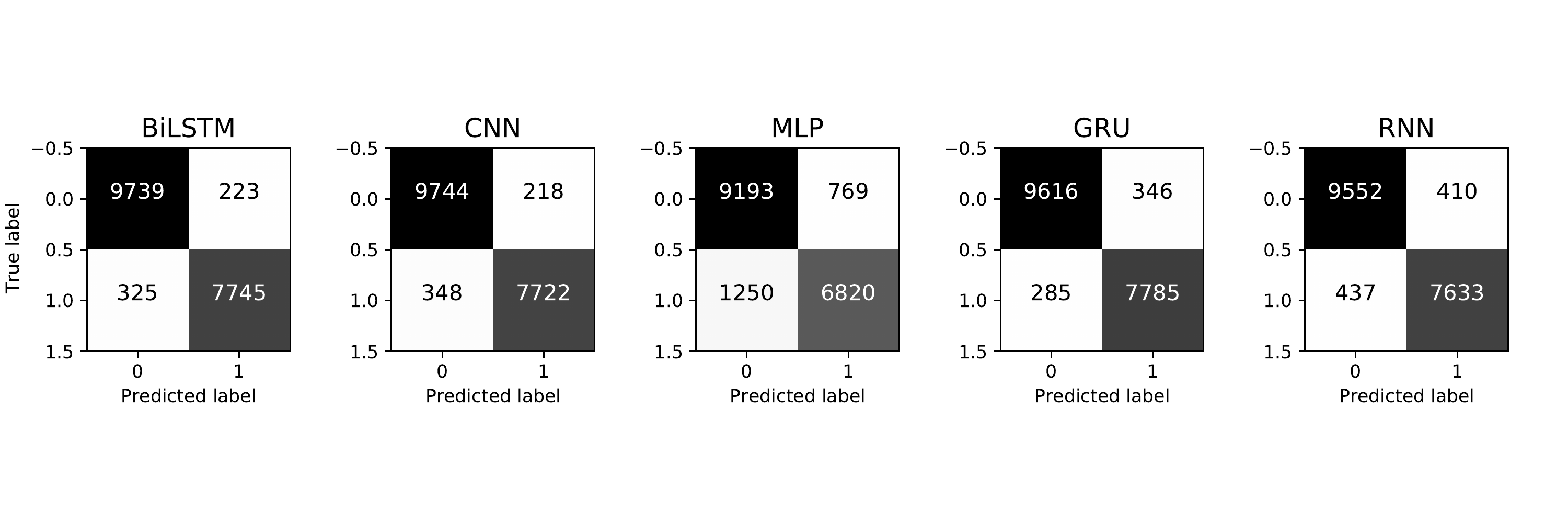}}    
\subfigure[]{   \includegraphics[width=1\linewidth]{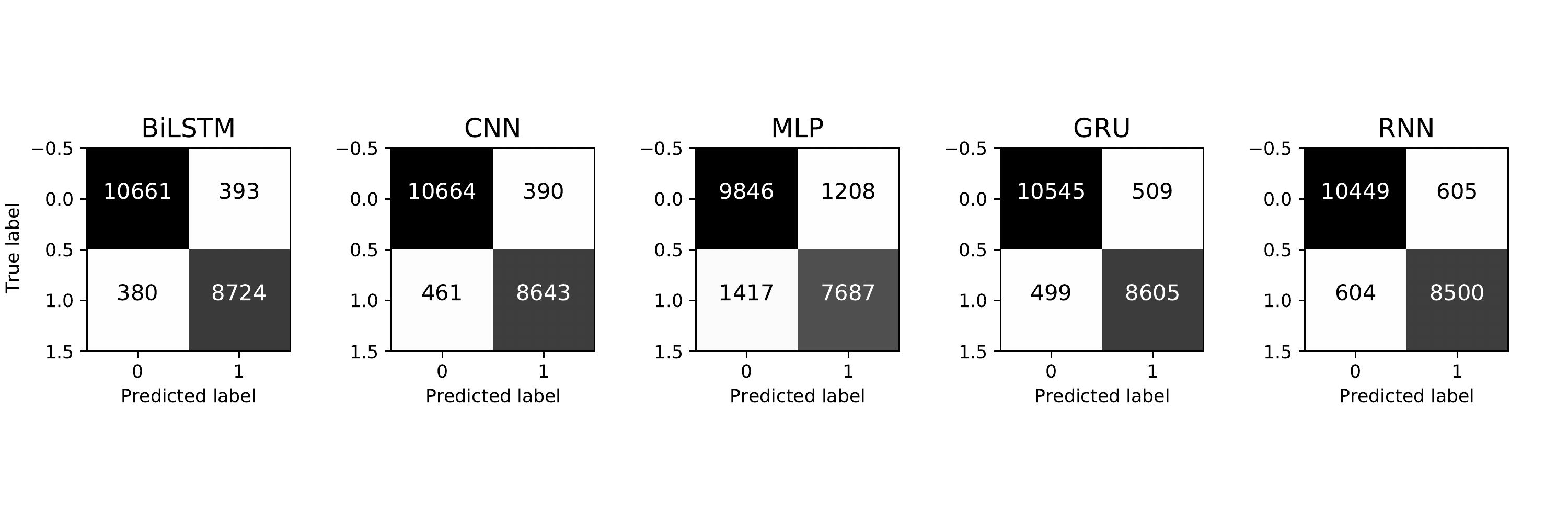}}
    \caption{Confusion matrices: (a) Dataset with only $100\%$ gender ratio names and (b) Dataset with all names regardless the gender ratio.}
    \label{matrix}
\end{figure}

For evaluating the approaches, we used the confusion matrix to measure the performance of deep learning models, as shown in Figures \ref{matrix} (a) and (b). Interpreting the confusion matrix, we have the predicted labels on the x-axis and the real labels on the y-axis. The black and gray cells carry the number of names that the models accurately predicted, and the white cells contain the number of names that the models incorrectly predicted. 
Analyzing the confusion matrix, we can conclude that the recurrent models RNN, BiLSTM, and GRU performed more favorable outcomes. In other words, fewer confusions occur in the classification with these models. Indeed, this is what accuracy implies.  But since accuracy is not regularly an adequate performance measure, we calculated the precision and recall, which is the true positive rate. The precision and recall are depicted in Tables \ref{tb} and \ref{t2}. By adopting the encoding word process previously described, all the models achieve considerable results, as shown by recall values. However, BiLSTM model with a precision of 0.9720 for the dataset with only $100\%$ gender ratio names and 0.9569 for the dataset with all names regardless the gender ratio, outperforms all other implemented models.

\section{Conclusion}

We presented how deep neural network models can process and classify Brazilian names using word encoding, more specifically character encoding. The deep network models presented are capable of successfully capture dependencies in the word vector. Moreover, we compared their performance with some common machine learning algorithms, such as SVM, KNN, and naive Bayes, extra tree, random forest, gradient boosting, logistic regression, ridge classifier, decision tree, Ada boost, LDA, and QDA. The extra tree and random forest classifier presented good performances. However, deep models such as CNN and BiLSTM showed better outcomes.  Through the validation, we concluded that the MLP model is suitable for classification prediction. However, during the comparison, the other models may appear more suited to the gender prediction problem. The benefit of using CNNs is their capacity to develop an internal representation. Nevertheless, because of the ability to deal with sequence prediction, recurrent models, such as BiLSTM achieved better results.

\bibliographystyle{IEEEtran}
\bibliography{ref}

\begin{thebibliography}{10}
\providecommand{\url}[1]{#1}
\csname url@samestyle\endcsname
\providecommand{\newblock}{\relax}
\providecommand{\bibinfo}[2]{#2}
\providecommand{\BIBentrySTDinterwordspacing}{\spaceskip=0pt\relax}
\providecommand{\BIBentryALTinterwordstretchfactor}{4}
\providecommand{\BIBentryALTinterwordspacing}{\spaceskip=\fontdimen2\font plus
\BIBentryALTinterwordstretchfactor\fontdimen3\font minus
  \fontdimen4\font\relax}
\providecommand{\BIBforeignlanguage}[2]{{%
\expandafter\ifx\csname l@#1\endcsname\relax
\typeout{** WARNING: IEEEtran.bst: No hyphenation pattern has been}%
\typeout{** loaded for the language `#1'. Using the pattern for}%
\typeout{** the default language instead.}%
\else
\language=\csname l@#1\endcsname
\fi
#2}}
\providecommand{\BIBdecl}{\relax}
\BIBdecl

\bibitem{karimi2016inferring}
F.~Karimi, C.~Wagner, F.~Lemmerich, M.~Jadidi, and M.~Strohmaier, ``Inferring
  gender from names on the web: A comparative evaluation of gender detection
  methods,'' in \emph{Proceedings of the 25th International conference
  companion on World Wide Web}, 2016, pp. 53--54.

\bibitem{vicente2019gender}
M.~Vicente, F.~Batista, and J.~P. Carvalho, ``Gender detection of twitter users
  based on multiple information sources,'' in \emph{Interactions Between
  Computational Intelligence and Mathematics Part 2}.\hskip 1em plus 0.5em
  minus 0.4em\relax Springer, 2019, pp. 39--54.

\bibitem{al2019study}
A.~I. Al-Ghadir and A.~M. Azmi, ``A study of arabic social media
  users—posting behavior and author’s gender prediction,'' \emph{Cognitive
  Computation}, vol.~11, no.~1, pp. 71--86, 2019.

\bibitem{ugail2018secrets}
H.~Ugail, ``Secrets of a smile? your gender and perhaps your biometric
  identity,'' \emph{Biometric Technology Today}, vol. 2018, no.~6, pp. 5--7,
  2018.

\bibitem{galla2020support}
D.~K.~K. Galla, B.~R. Mukamalla, and R.~P.~R. Chegireddy, ``Support vector
  machine based feature extraction for gender recognition from objects using
  lasso classifier,'' \emph{Journal of Big Data}, vol.~7, no.~1, pp. 1--16,
  2020.

\bibitem{afifi2019afif4}
M.~Afifi and A.~Abdelhamed, ``Afif4: deep gender classification based on
  adaboost-based fusion of isolated facial features and foggy faces,''
  \emph{Journal of Visual Communication and Image Representation}, vol.~62, pp.
  77--86, 2019.

\bibitem{panchenko2014detecting}
A.~Panchenko and A.~Teterin, ``Detecting gender by full name: Experiments with
  the russian language,'' in \emph{International Conference on Analysis of
  Images, Social Networks and Texts}.\hskip 1em plus 0.5em minus 0.4em\relax
  Springer, 2014, pp. 169--182.

\bibitem{van2020gender}
J.~van~de Weijer, G.~Ren, J.~van~de Weijer, W.~Wei, and Y.~Wang, ``Gender
  identification in chinese names,'' \emph{Lingua}, vol. 234, p. 102759, 2020.

\bibitem{hu2021s}
Y.~Hu, C.~Hu, T.~Tran, T.~Kasturi, E.~Joseph, and M.~Gillingham, ``What’s in
  a name?--gender classification of names with character based machine learning
  models,'' \emph{Data Mining and Knowledge Discovery}, pp. 1--27, 2021.

\bibitem{Age2017}
K.~Zhang, C.~Gao, L.~Guo, M.~Sun, X.~Yuan, T.~X. Han, Z.~Zhao, and B.~Li, ``Age
  group and gender estimation in the wild with deep ror architecture,''
  \emph{IEEE Access}, vol.~5, pp. 22\,492--22\,503, 2017.

\bibitem{transfer2020}
C.~H. Nga, K.-T. Nguyen, N.~C. Tran, and J.-C. Wang, ``Transfer learning for
  gender and age prediction,'' in \emph{2020 IEEE International Conference on
  Consumer Electronics - Taiwan (ICCE-Taiwan)}, 2020, pp. 1--2.

\bibitem{SVM2020}
A.~Venugopal, O.~Yadukrishnan, and R.~Nair~T., ``A svm based gender
  classification from children facial images using local binary and non-binary
  descriptors,'' in \emph{2020 Fourth International Conference on Computing
  Methodologies and Communication (ICCMC)}, 2020, pp. 631--634.

\bibitem{Census2020}
S.~Mittal and V.~S. Rajput, ``Gender and age based census system for
  metropolitan cities,'' in \emph{2020 8th International Conference on
  Reliability, Infocom Technologies and Optimization (Trends and Future
  Directions) (ICRITO)}, 2020, pp. 1094--1097.

\bibitem{predict2019}
A.~Kuehlkamp and K.~Bowyer, ``Predicting gender from iris texture may be harder
  than it seems,'' in \emph{2019 IEEE Winter Conference on Applications of
  Computer Vision (WACV)}, 2019, pp. 904--912.

\bibitem{surinta2019}
O.~Surinta and T.~Khamket, ``Gender recognition from facial images using local
  gradient feature descriptors,'' in \emph{2019 14th International Joint
  Symposium on Artificial Intelligence and Natural Language Processing
  (iSAI-NLP)}, 2019, pp. 1--6.

\bibitem{Vashisth2020}
P.~Vashisth and K.~Meehan, ``Gender classification using twitter text data,''
  in \emph{2020 31st Irish Signals and Systems Conference (ISSC)}, 2020, pp.
  1--6.

\bibitem{Lekamge2019}
T.~Lekamge and T.~Fernando, ``Finding the gender of personal names and finding
  the effect of gana on personal names with long short term memory,'' in
  \emph{2019 19th International Conference on Advances in ICT for Emerging
  Regions (ICTer)}, vol. 250, 2019, pp. 1--8.

\bibitem{Mamgain2019}
S.~Mamgain, R.~C~Balabantaray, and A.~K~Das, ``Author profiling: Prediction of
  gender and language variety from document,'' in \emph{2019 International
  Conference on Information Technology (ICIT)}, 2019, pp. 473--477.

\bibitem{Otter2021}
D.~W. Otter, J.~R. Medina, and J.~K. Kalita, ``A survey of the usages of deep
  learning for natural language processing,'' \emph{IEEE Transactions on Neural
  Networks and Learning Systems}, vol.~32, no.~2, pp. 604--624, 2021.

\bibitem{goodfellow2016deep}
I.~Goodfellow, Y.~Bengio, A.~Courville, and Y.~Bengio, \emph{Deep
  learning}.\hskip 1em plus 0.5em minus 0.4em\relax MIT press Cambridge, 2016,
  vol.~1.

\bibitem{Amarappa2015Kannada}
S.~Amarappa and S.~Sathyanarayana, ``Kannada named entity recognition and
  classification (nerc) based on multinomial naïve bayes (mnb) classifier,''
  \emph{International Journal on Natural Language Computing}, vol.~4, 09 2015.

\bibitem{manik2019gender}
L.~P. Manik, A.~F. Syafiandini, H.~F. Mustika, Z.~Akbar, and Y.~Rianto,
  ``Gender inference based on indonesian name and profile photo,'' in
  \emph{2019 International Conference on Computer, Control, Informatics and its
  Applications (IC3INA)}.\hskip 1em plus 0.5em minus 0.4em\relax IEEE, 2019,
  pp. 25--29.

\bibitem{yuenyong2020gender}
S.~Yuenyong and S.~Sinthupinyo, ``Gender classification of thai facebook
  usernames,'' \emph{International Journal of Machine Learning and Computing},
  vol.~10, no.~5, 2020.

\bibitem{mikolov2013efficient}
T.~Mikolov, K.~Chen, G.~Corrado, and J.~Dean, ``Efficient estimation of word
  representations in vector space,'' \emph{arXiv preprint arXiv:1301.3781},
  2013.

\bibitem{quinlan1986induction}
J.~R. Quinlan, ``Induction of decision trees,'' \emph{Machine learning},
  vol.~1, no.~1, pp. 81--106, 1986.

\bibitem{potharst2000decision}
R.~Potharst and J.~C. Bioch, ``Decision trees for ordinal classification,''
  \emph{Intelligent Data Analysis}, vol.~4, no.~2, pp. 97--111, 2000.

\bibitem{ahmad2018performance}
I.~Ahmad, M.~Basheri, M.~J. Iqbal, and A.~Rahim, ``Performance comparison of
  support vector machine, random forest, and extreme learning machine for
  intrusion detection,'' \emph{IEEE access}, vol.~6, pp. 33\,789--33\,795,
  2018.

\bibitem{alsariera2020ai}
Y.~A. Alsariera, V.~E. Adeyemo, A.~O. Balogun, and A.~K. Alazzawi, ``Ai
  meta-learners and extra-trees algorithm for the detection of phishing
  websites,'' \emph{IEEE Access}, vol.~8, pp. 142\,532--142\,542, 2020.

\bibitem{alzamzami2020light}
F.~Alzamzami, M.~Hoda, and A.~El~Saddik, ``Light gradient boosting machine for
  general sentiment classification on short texts: A comparative evaluation,''
  \emph{IEEE Access}, vol.~8, pp. 101\,840--101\,858, 2020.

\bibitem{jan2017sensor}
S.~U. Jan, Y.-D. Lee, J.~Shin, and I.~Koo, ``Sensor fault classification based
  on support vector machine and statistical time-domain features,'' \emph{IEEE
  Access}, vol.~5, pp. 8682--8690, 2017.

\bibitem{xue2020real}
Z.~Xue, J.~Wei, and W.~Guo, ``A real-time naive bayes classifier accelerator on
  fpga,'' \emph{IEEE Access}, vol.~8, pp. 40\,755--40\,766, 2020.

\bibitem{fix1989discriminatory}
E.~Fix and J.~L. Hodges, ``Discriminatory analysis. nonparametric
  discrimination: Consistency properties,'' \emph{International Statistical
  Review/Revue Internationale de Statistique}, vol.~57, no.~3, pp. 238--247,
  1989.

\bibitem{zhao2018knn}
F.~Zhao and Q.~Tang, ``A knn learning algorithm for collusion-resistant
  spectrum auction in small cell networks,'' \emph{IEEE Access}, vol.~6, pp.
  45\,796--45\,803, 2018.

\bibitem{hong2017comparative}
H.~Hong, S.~A. Naghibi, M.~M. Dashtpagerdi, H.~R. Pourghasemi, and W.~Chen, ``A
  comparative assessment between linear and quadratic discriminant analyses
  (lda-qda) with frequency ratio and weights-of-evidence models for forest fire
  susceptibility mapping in china,'' \emph{Arabian Journal of Geosciences},
  vol.~10, no.~7, p. 167, 2017.

\bibitem{kamath2019deep}
U.~Kamath, J.~Liu, and J.~Whitaker, \emph{Deep learning for NLP and speech
  recognition}.\hskip 1em plus 0.5em minus 0.4em\relax Springer, 2019, vol.~84.

\bibitem{zhang2015character}
X.~Zhang, J.~Zhao, and Y.~Lecun, ``Character-level convolutional networks for
  text classification,'' \emph{Advances in Neural Information Processing
  Systems}, vol. 2015, pp. 649--657, 2015.

\bibitem{haykin2010neural}
S.~Haykin, \emph{Neural Networks and Learning Machines, 3/E}.\hskip 1em plus
  0.5em minus 0.4em\relax Pearson Education India, 2010.

\bibitem{hochreiter1997long}
S.~Hochreiter and J.~Schmidhuber, ``Long short-term memory,'' \emph{Neural
  computation}, vol.~9, no.~8, pp. 1735--1780, 1997.

\bibitem{cho2014learning}
K.~Cho, B.~Van~Merri{\"e}nboer, C.~Gulcehre, D.~Bahdanau, F.~Bougares,
  H.~Schwenk, and Y.~Bengio, ``Learning phrase representations using rnn
  encoder-decoder for statistical machine translation,'' \emph{arXiv preprint
  arXiv:1406.1078}, 2014.

\bibitem{chung2014empirical}
J.~Chung, C.~Gulcehre, K.~Cho, and Y.~Bengio, ``Empirical evaluation of gated
  recurrent neural networks on sequence modeling,'' in \emph{NIPS 2014 Workshop
  on Deep Learning, December 2014}, 2014.

\end{thebibliography}

\end{document}